\documentclass[10pt,twocolumn,letterpaper]{article}

\usepackage{wacv}
\usepackage{times}
\usepackage{epsfig}
\usepackage{graphicx}
\usepackage{amsmath}
\usepackage{amssymb}

\wacvfinalcopy

\begin{document}

\title{COCO-Text: Dataset and Benchmark for Text Detection and Recognition in Natural Images}

\author{
Andreas Veit\textsuperscript{1,2}, 
Tom{\'a}{\v s} Matera\textsuperscript{2}, 
Luk{\'a}{\v s} Neumann\textsuperscript{3}, 
Ji{\v r}{\'i} Matas\textsuperscript{3}, 
Serge Belongie\textsuperscript{1,2}\\
\\
\textsuperscript{1} Department of Computer Science, Cornell University \quad \textsuperscript{2} Cornell Tech\\
\textsuperscript{3} Department of Cybernetics, Czech Technical University, Prague\\
{\tt\small \textsuperscript{1}\{av443,sjb344\}@cornell.edu, \textsuperscript{2}tomas@matera.cz,  \textsuperscript{3}\{neumalu1,matas\}@cmp.felk.cvut.cz}\\
}

\maketitle
\ifwacvfinal\thispagestyle{empty}\fi

\begin{abstract}
This paper describes the COCO-Text dataset. In recent years large-scale datasets like SUN and Imagenet drove the advancement of scene understanding and object recognition. The goal of COCO-Text is to advance state-of-the-art in text detection and recognition in natural images. The dataset is based on the MS COCO dataset, which contains images of complex everyday scenes. The images were not collected with text in mind and thus contain a broad variety of text instances. To reflect the diversity of text in natural scenes, we annotate text with (a) location in terms of a bounding box, (b) fine-grained classification into machine printed text and handwritten text, (c) classification into legible and illegible text, (d) script of the text and (e) transcriptions of legible text. The dataset contains over 173k text annotations in over 63k images. We provide a statistical analysis of the accuracy of our annotations.
In addition, we present an analysis of three leading state-of-the-art photo Optical Character Recognition (OCR) approaches on our dataset. While scene text detection and recognition enjoys strong advances in recent years, we identify significant shortcomings motivating future work. 
\end{abstract}

\vspace{-6pt}
\section{Introduction}
The detection and recognition of scene text in natural images with unconstrained environments remains a challenging problem in computer vision. The ability to robustly read text in unconstrained scenes can significantly help with numerous real-world application, \eg, assistive technology for the visually impaired, robot navigation and geo-localization. The problem of detecting and recognizing text in scenes has received increasing attention from the computer vision community in recent years~\cite{bissacco2013photoocr,jaderberg2014reading,neumann2012real,wang2011end}. The evaluation is typically done on datasets comprising images with mostly iconic text and containing hundreds of images at best.

\begin{figure}[t]
\includegraphics[width=1\linewidth]{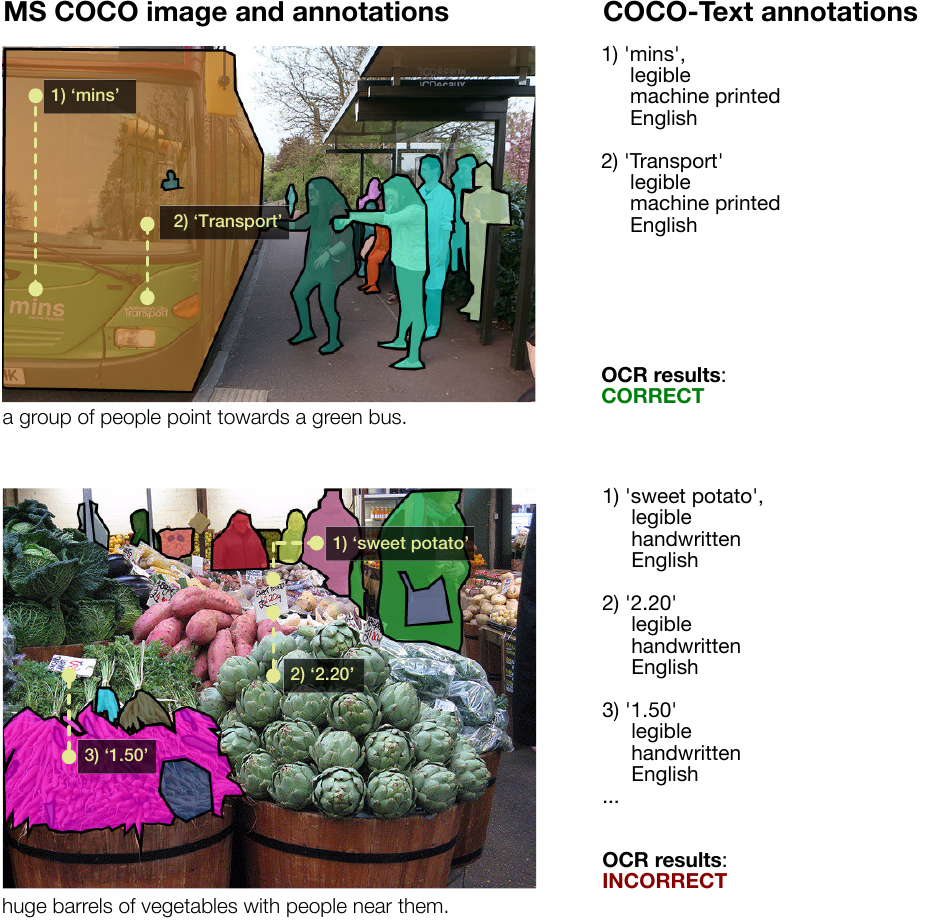}
\caption{
Left: Example MS COCO images with object segmentation and captions. Right: COCO-Text annotations. For the top image, the photo OCR finds and recognizes the text printed on the bus. For the bottom image, the OCR does not recognize the handwritten price tags on the fruit stand. 
}
\label{fig:page1}
\vspace{-14pt}
\end{figure}

To advance the understanding of text in unconstrained scenes we present a new large-scale dataset for text in natural images. The dataset is based on the Microsoft COCO dataset~\cite{lin2014microsoft} that annotates common objects in their natural contexts. Combining rich text annotations and object annotations in natural images provides a great opportunity for research in scene text detection and recognition. MS COCO was not collected with text in mind and is thus potentially less biased towards text. Further, combining text with object annotations allows for contextual reasoning about scene text and objects. During a pilot study of state-of-the-art photo OCR methods on MS COCO, we made two key observations: First, text in natural scenes is very diverse ranging from legible machine printed text on street signs to illegible graffiti and handwritten notes. Second, while the domain of scene text detection and recognition enjoys significant advances in recent years, there is still way to go to reach the performance needed for real world applications. Figure~\ref{fig:page1} shows sample images from the dataset illustrating the diversity of scene text in natural images and the challenges for text detection and recognition.

The main contributions of this work is the COCO-Text dataset.\footnote{available at http://vision.cornell.edu/se3/coco-text} The purpose of the dataset is to provide the research community with a resource to advance the state-of-the-art in scene text detection and recognition as well as help evaluating shortcomings of existing methods.
The dataset contains 63,686 images with 173,589 labeled text regions. For each text region, we provide the location in terms of bounding boxes, classifications in terms of legibility, category (\eg machine printed or hand written) and script of the text, as well as transcriptions in case of legible text with western script. We also provide a detailed description of our annotation pipeline. In addition, we present an analysis of three leading state-of-the-art photo OCR algorithms on our dataset. The results show that some methods achieve excellent detection precision and good transcription accuracy. However, recall for text detection is considerably degraded. In particular, for illegible text none of the methods shows viable functionality. These significant shortcomings motivate future work.

\section{Related Work}
In this paper, we introduce a large-scale dataset for text in natural images to support the advancement of data driven scene text detection and recognition methods. Therefore, we restrict this discussion to related datasets, state-of-the-art scene text recognition, approaches combining text with context cues and advances in labeling task assignments.

\begin{figure}[h]
\begin{center}
\begin{tabular}{@{}c@{}c@{}}
\includegraphics[width=\linewidth]{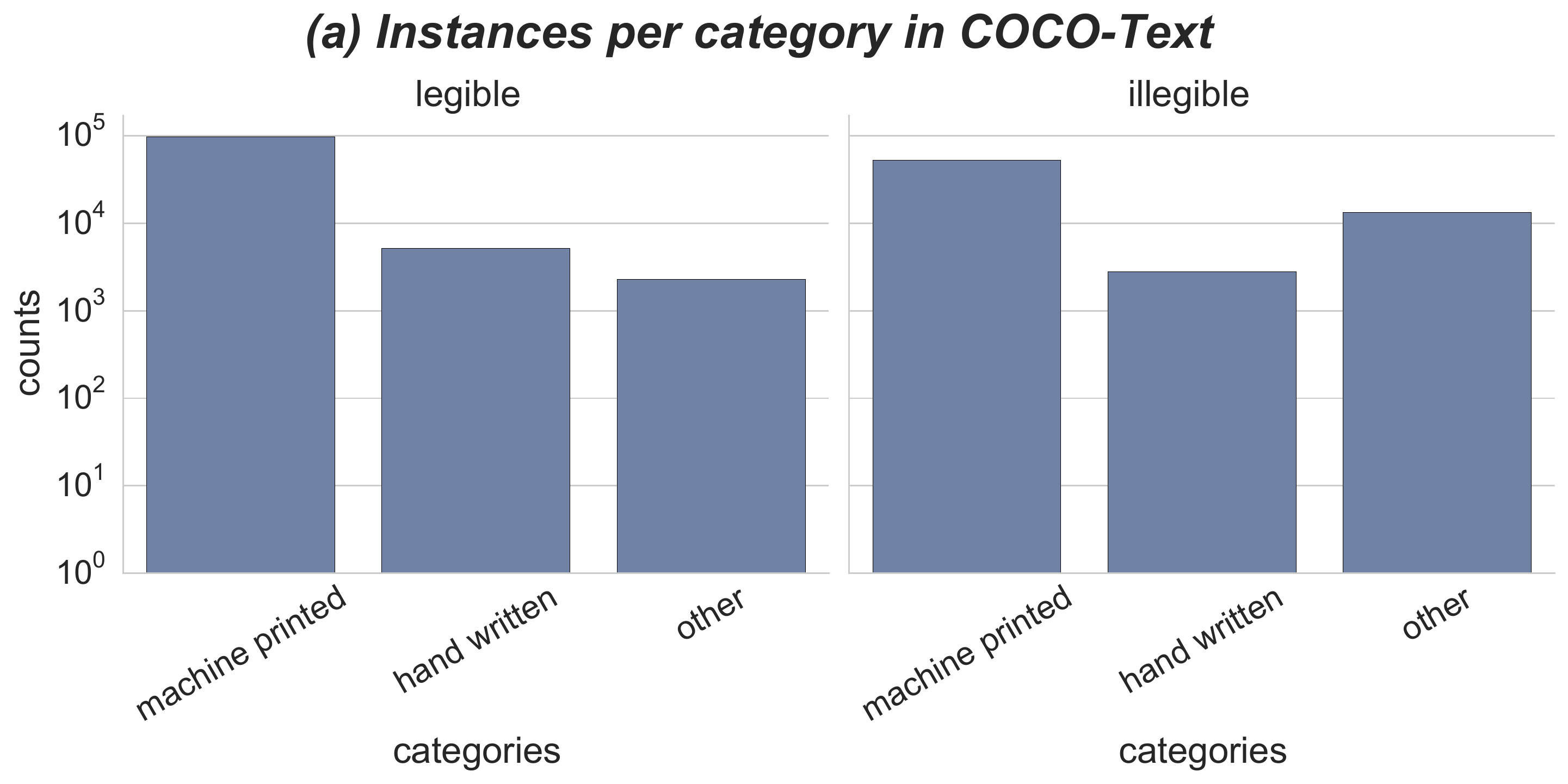} \\
\includegraphics[width=\linewidth]{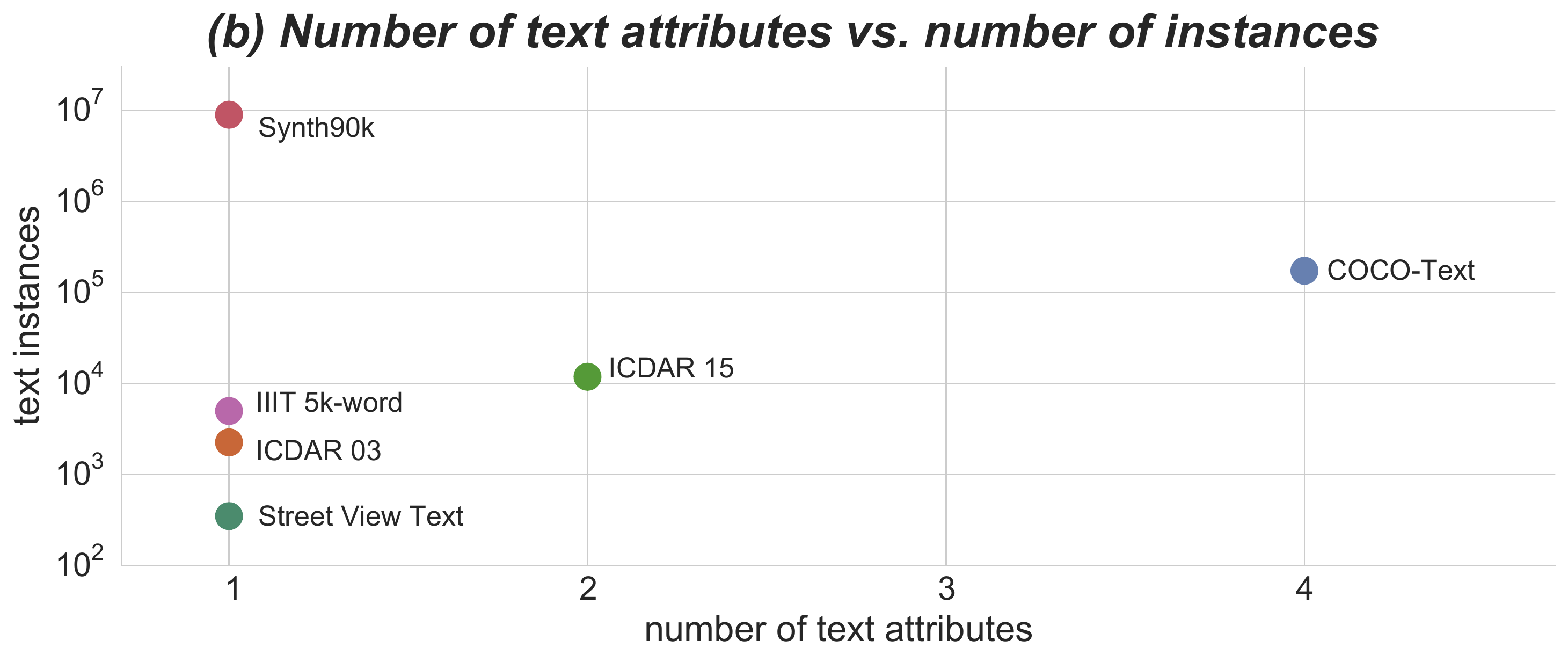} \\
\includegraphics[width=\linewidth]{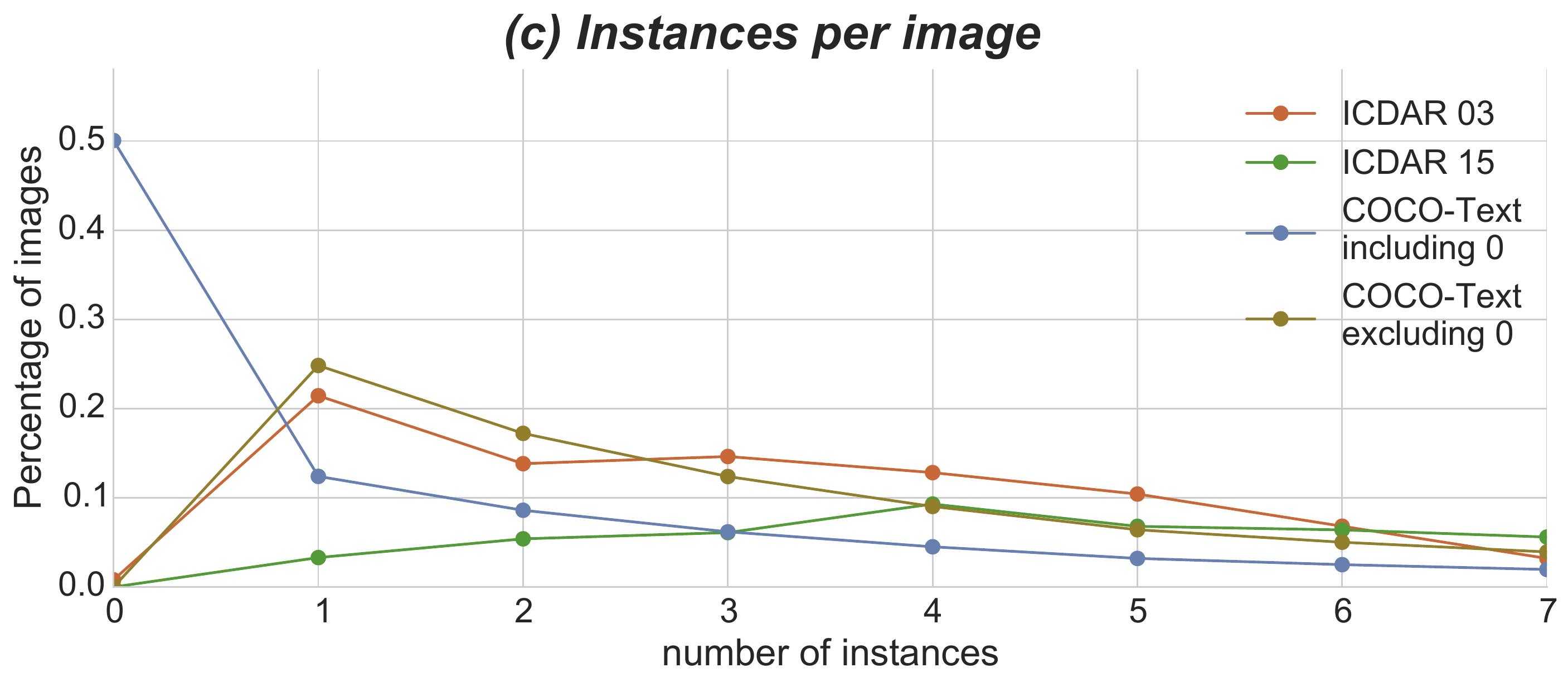} \\
\includegraphics[width=\linewidth]{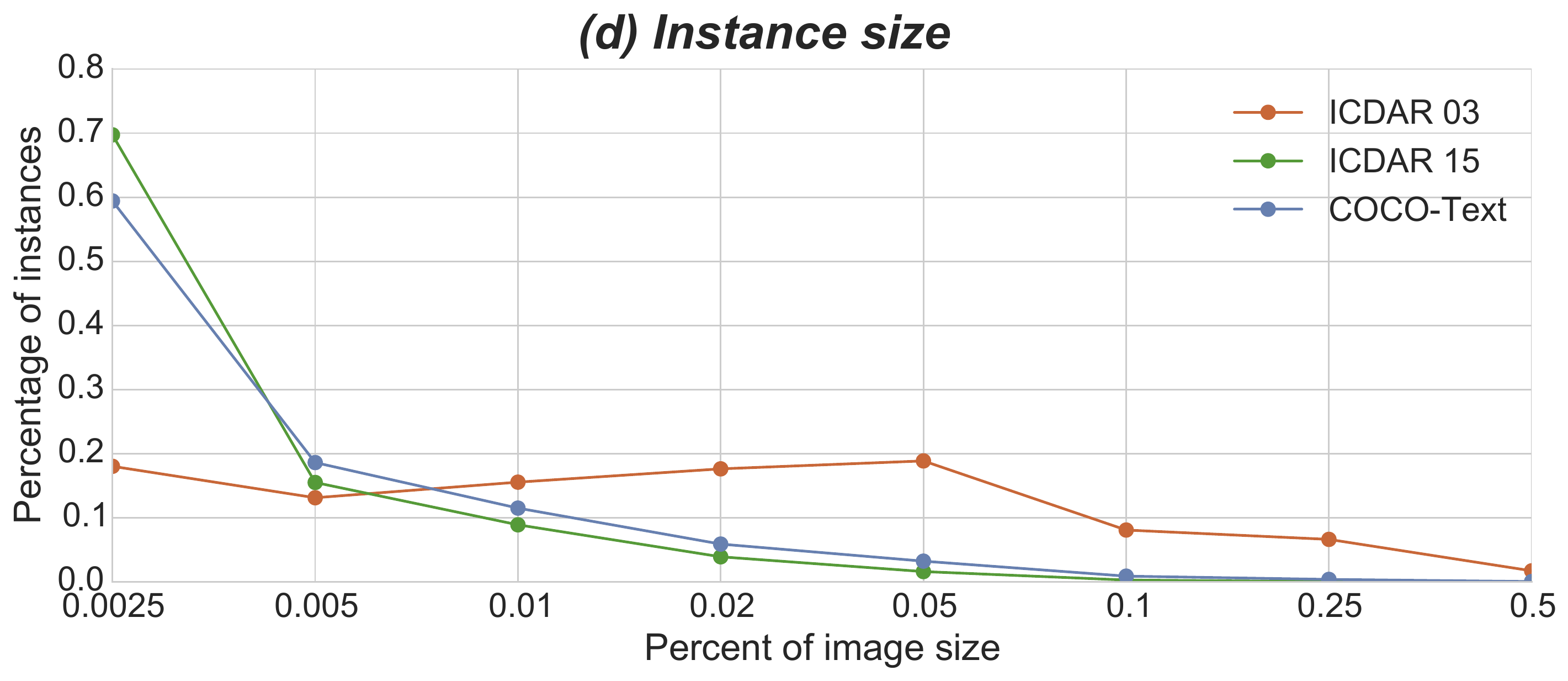} \\
\includegraphics[width=\linewidth]{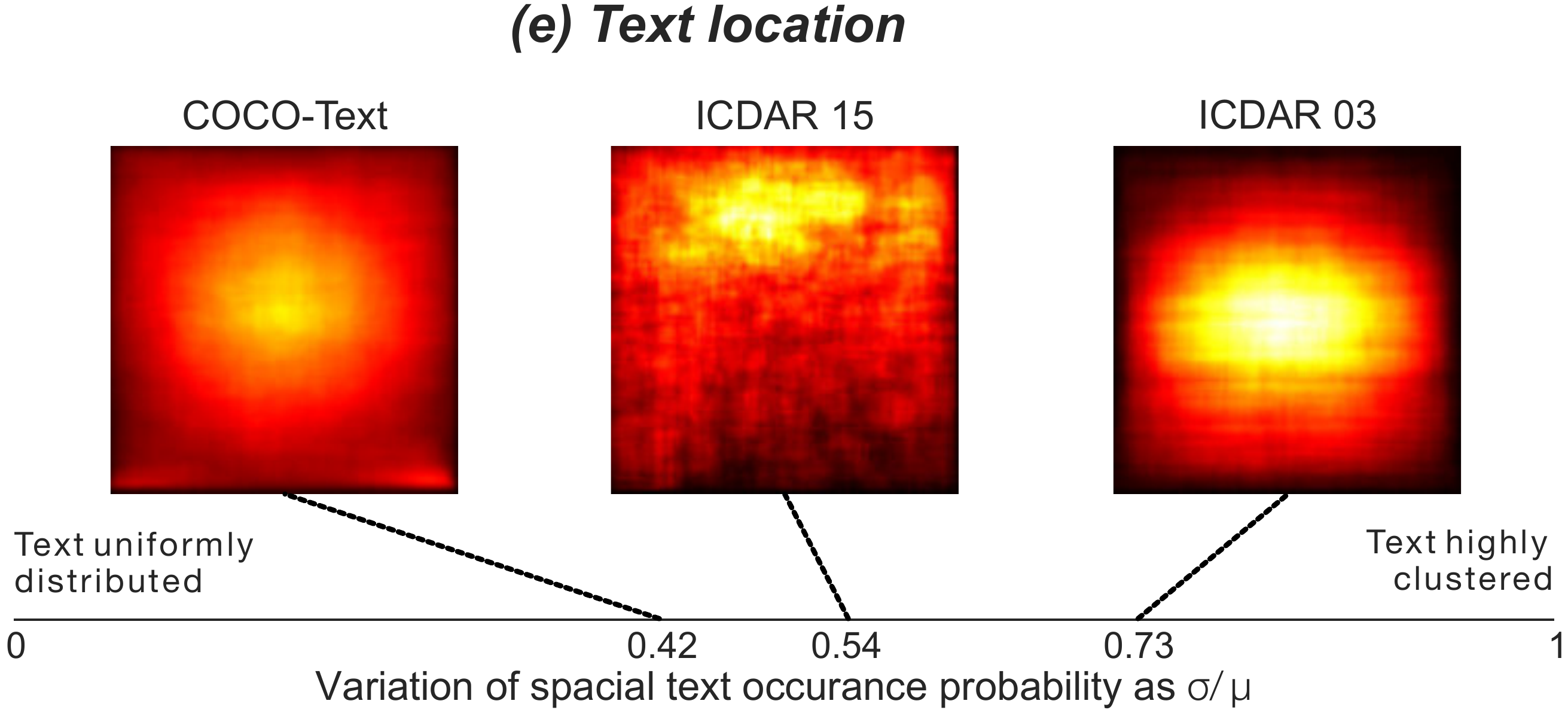} \\
\end{tabular}
\end{center}
\vspace{-4pt}
\caption{\textbf{Dataset statistics:} (a) Number of annotated instances per category for COCO-Text. (b) Number of text attributes vs the number of instances for popular scene text datasets. (c) Number of annotated instances per image for COCO-Text, ICDAR03 and ICDAR15. (d) Distribution of instance sizes for same datasets. (e) Spacial text occurrence probability for same datasets.}
\label{fig:datasets}
\vspace{-14pt}
\end{figure}

In recent years large-scale datasets like SUN \cite{xiao2010sun}, Imagenet \cite{deng2009imagenet} and MS COCO \cite{lin2014microsoft} drove the advancement of several fields in computer vision. The presented dataset is based upon MS COCO  and its image captions extension \cite{chen2015microsoft}. We utilize the rich annotations from these datasets to optimize annotators' task allocations. 

This work lies in the context of other scene text datasets. Figure~\ref{fig:datasets} compares COCO-Text to related datasets. The ICDAR Robust Reading challenge \cite{lucas2003icdar}, referred to from now on as ICDAR 03, was the first public dataset for detecting and recognizing scene text. The dataset contains 509 scene images and the scene text is mostly centered and iconic. The newest iteration of this challenge \cite{karatzas2015icdar} introduced a competition on \emph{incidental scene text}, referred to as ICDAR 15, which contains about 1500 images acquired with wearable devices. These images contain 11,886 text instances. Further, Street View Text \cite{wang2010word} contains a total of 350 images from Google Street View and 725 total labeled words, however it only contains annotations for a fraction of text in the images. Other scene text datasets include IIIT 5k-word \cite{mishra2012scene} which contains 3000 cropped word images of scene text downloaded from Google image search as well as Synth90k \cite{jaderberg2014synthetic}, a dataset of 9 million cropped word images that have been synthetically generated to train systems for character recognition. The presented COCO-Text dataset differs from the previous datasets in three key aspects: First, images in MS COCO were not selected with text in mind. Thus, the annotated text instances lie in their natural context. This is particularly important for text detection in complex everyday scenes. Further, the text has a broader distribution of spatial occurrence Figure~\ref{fig:datasets}~(e). Second, COCO-Text contains a wide variety of text instances and we annotate fine-grained categories such as machine printed text and handwritten text, text legibility as well as the script. Third, COCO-Text has a much larger scale than other datasets for text detection and recognition. In particular, our dataset has over 14 times more text annotations than related datasets. Figure~\ref{fig:datasets}~(b) gives an overview of the dataset sizes and number of annotated attributes.

Scene text detection and recognition approaches generally comprise two parts: Detecting proposal text regions in the image, and recognizing the words in those regions. Current work in the area include approaches by Bissacco et al.~\cite{bissacco2013photoocr}, where first three different detectors are combined to identify text regions and subsequently characters are classified with a fully connected neural network with HOG features as input supported by a language model based on n-grams. Further, Neumann and Matas~\cite{neumann2012real} first identify Extremal Regions, groups them into words and then selects most probable character segmentation. Furthermore, Jaderberg et al.~\cite{jaderberg2014reading} use Convolutional Neural Networks (CNN) for both text region detection and character classification. 

Further related work that recently receives more attention is combining textual and visual cues for fine-grained image classification. For example, Rusi{\~n}ol et al.~\cite{rusinol2014multimodal} merge visual and textual descriptions to classify administrative document images and Karaoglu et al.~\cite{Karaoglu13} use detected scene text for fine-grained building classification.

Another related stream of research focuses on repeated labeling in the face of noisy labels as well as combining human workers with machine classifiers. In early work, Romney et al.~\cite{romney1986culture} looked into improving label quality by taking worker accuracy into account. Further, Wilber at al.~\cite{wilber2014cost} investigate the use of grid questions, where workers select answers from a grid of images to take advantage of the parallelism of human perception. We also use similar grid interfaces, but our approach differs in that we do not require a specific number of responses, because we perform binary classification whereas they do relative comparisons. Closer to our work, Russakovsky et al.~\cite{RussakovskyCVPR15} propose a framework that combines object detectors with human annotators to annotate a dataset. While we also combine results from object detectors and human annotators, our work differs in that we do not have access to the detectors during the annotation process, but only initial detections as input. In this area, the work closest to ours is the approach by Veit et al.~\cite{veit2015optimizing} that proposes strategies to optimize the task allocation to human workers with constrained budgets. We adapt their approach in that we increase annotation redundancy only the supposedly most difficult annotations.

\begin{figure}[t]
	\begin{center}
		\begin{tabular}{@{}c@{}c@{}}
			\includegraphics[width=0.8\linewidth]{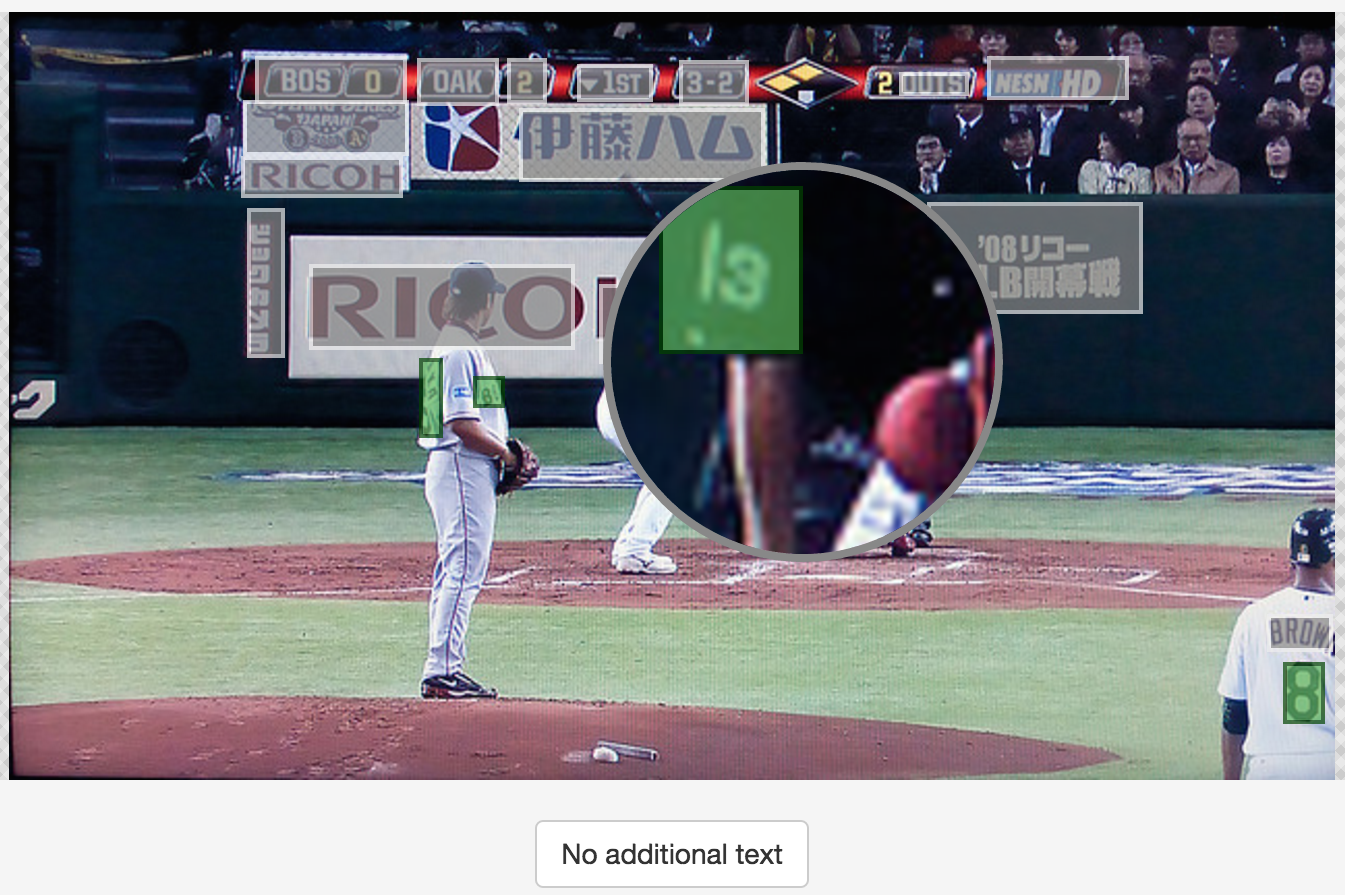}  \\
			(a) Text Spotting UI \\
			\includegraphics[width=0.8\linewidth]{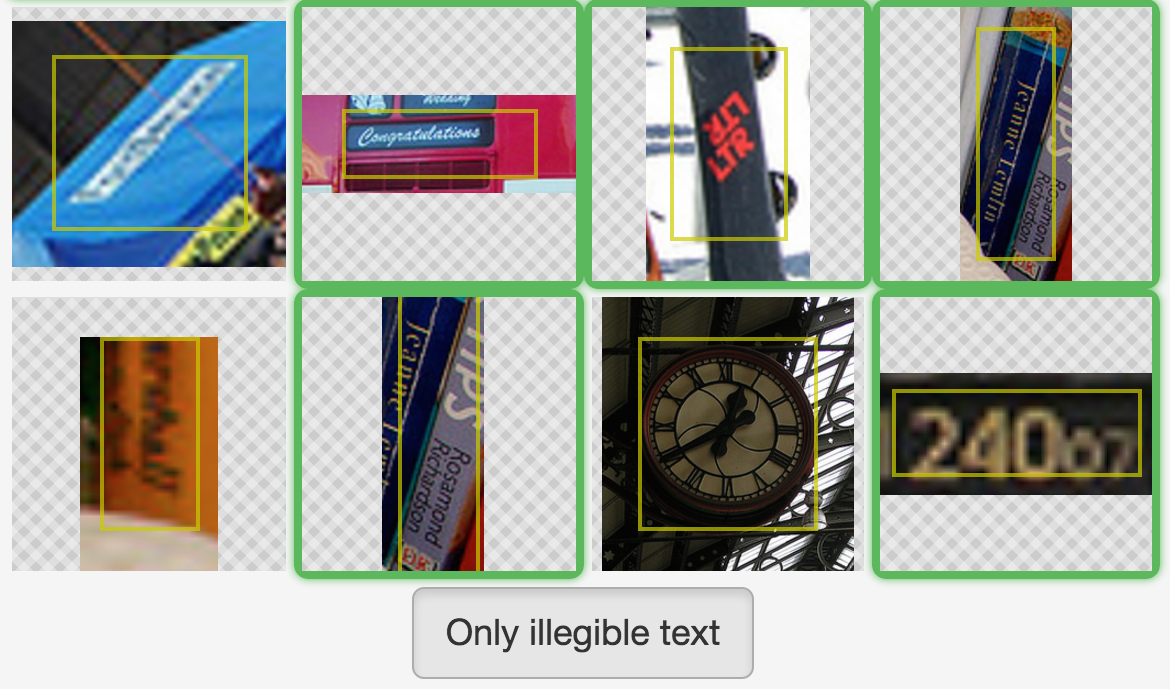} \\
			(b) Text Classification UI \\
			\includegraphics[width=0.8\linewidth]{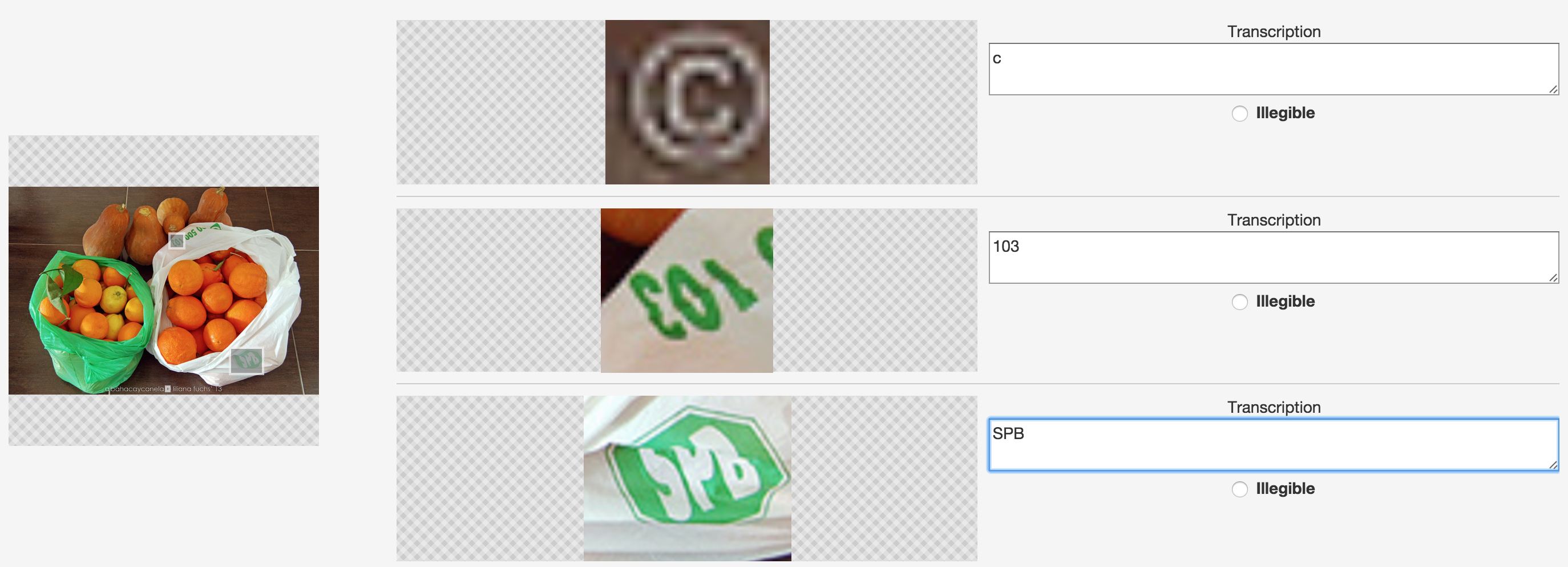} \\
			(c) Transcription UI
		\end{tabular}
	\end{center}
	\vspace{0pt}
	\caption{(a) In the text spotting UI, annotators can draw bounding boxes around text regions. Text spotted in previous rounds is marked with gray boxes. Worker can use a magnifying class for small text regions. (b) In the text classification UI, annotators can select text regions with a certain property. In the shown example, legible text is selected. The patches show a crop of the respective MS COCO images with a 20\% padding around the box. (c) In the transcription UI, annotators can transcribe presented text regions. If they can't read the text, they can flag it as illegible.}
	\label{fig:pipeline}
	\vspace{-16pt}
\end{figure}

\section{Text Annotation}
\begin{figure}[t]
	\includegraphics[width=1\linewidth]{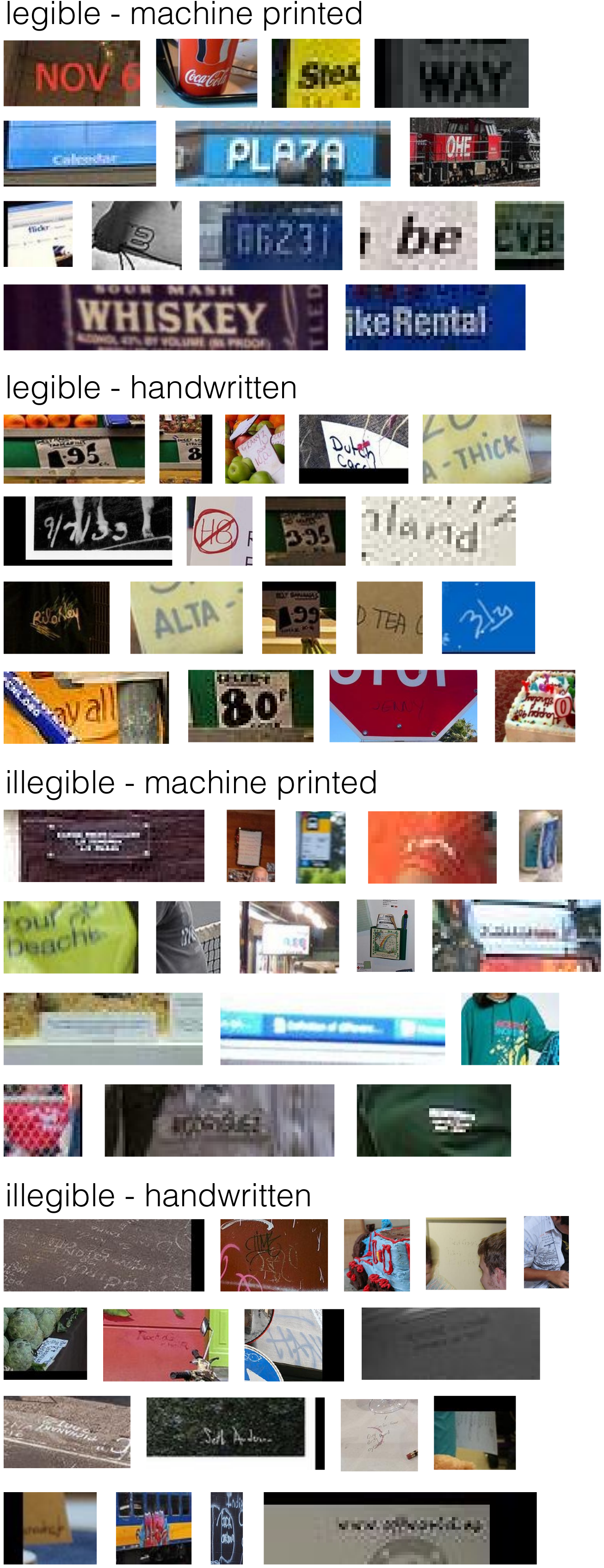}
	\caption{
	Crops around example text instances in COCO-Text organized by text categories.
	}
	\label{fig:lastfig}
	\vspace{-27pt}
\end{figure}

We now describe how we annotated the dataset. Due to our goal to annotate all text in the MS COCO dataset within a reasonable budget, the design of a cost efficient yet high quality annotation pipeline was essential. 
Table~\ref{tab:attributes} gives an overview of the annotations we collect for each text region. Crops around example text instances in COCO-Text organized by text categories are shown in Figure~\ref{fig:lastfig}

\begin{table}[t]
	\centering
	\caption{Annotations for each text region}
	\label{tab:attributes}
\begin{tabular}{l|l}
annotation & values \\
\hline 
 location	&  bounding box \\ 
 legibility	&  legible and illegible\\ 
 category  	&  machine printed, handwritten, others\\ 
 script 	&  English, not English\\ 
 transcription	&  utf8 string\\ 
\end{tabular} 
\vspace{-12pt}
\end{table}

\begin{figure}[t]
	\includegraphics[width=1\linewidth]{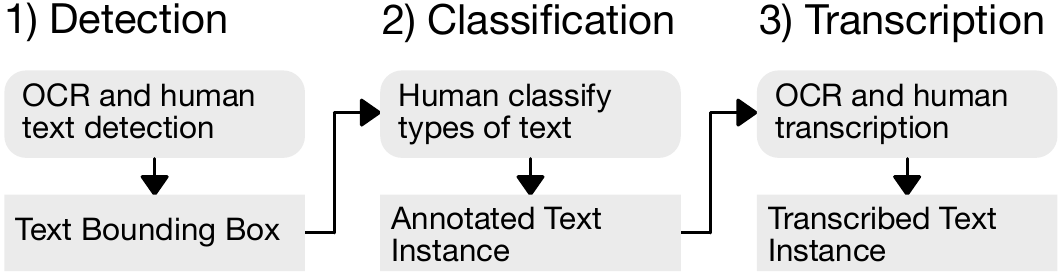}
	\caption{
		The pipeline has three steps: Detecting text regions, classifying them into fine-grained categories and transcription.
	}
	\label{fig:diagram}
	\vspace{-11pt}
\end{figure}

We use workers on Amazon's Mechanical Turk platform for all crowdsourcing tasks. To manage the different UIs and HITs we use the task manager from Matera et al.~\cite{matera2014user}. We keep the tasks as atomic as possible. For example, we separate the detection of text from deciding its category, because workers might determine with ease that no text is present in a scene without looking out for specific text. 
To reduce annotation time, we use the rich MS COCO annotations to help guide which images need more attention. Further, we received detection and transcription results from three leading state-of-the-art photo OCR algorithms. We are very grateful for the collaboration and use the results to reduce the workload of crowd workers. To make sure annotations are not biased towards particular OCR implementations, we make sure that each image is at least viewed and annotated by one human worker. The annotation pipeline is separated into three parts: Detection of text regions, classification of the text into fine grained categories and transcription. Figure~\ref{fig:diagram} gives an overview.

To ensure high annotation quality we employ four quality control mechanisms: First, we provide annotators with tutorials for each group of similar Human Intelligent Tasks (HITs) that describe the task and only let annotators start working on them once they successfully passed two examples. This ensures that they understand the task. In addition, we limit the number of images per worker to 150 to receive a good variety of answers. Further, we use \emph{catch trials}, where we know the ground truth to assess the quality of workers to post hoc decide whether to use their annotations. Lastly, we introduce \emph{stop and go trials} where we also know the ground truth and workers get immediate feedback for wrong answers. Specifically, in case of a wrong answer, workers have to wait for 5-10 seconds until they can continue and try again. These \emph{stop and go trials} proved particularly effective. The reason behind the effectiveness is that human annotators face a trade-off between answer quality and invested time. Introducing immediate time penalty for low quality answers, directly affects the trade-off such that workers provide higher quality answers. To ensure workers' satisfaction it is key that the examples are not ambiguous and the threshold is low enough so that workers with average quality answers do not get stopped.

\subsection{Text Region Detection}
The first task for annotating the dataset is detecting text regions present in each image. We annotate each text region with an enclosing bounding box. For legible text we aim for one bounding box per word, \ie an uninterrupted sequence of characters separated by a space, and for illegible text we aim for one bounding box per continuous text region, \eg a sheet or paper. The text detection step has four parts: 

\textbf{Incorporating Photo OCR input\ } 
First, we use the input provided by our collaborating photo OCR approaches. In particular, we use the detection results. We treat each detection as if it were a human annotator. False positive detections are handled in a subsequent stage after we collect the human detections. To reduce bias towards specific approaches, we only use OCR input, where at least one human annotator agrees with the OCR input. This step contributes about 20\% of our text annotations. 

\textbf{Spotting text not found by any OCR\ } 
Second, we ask human workers to annotate the remaining text regions. We ask workers to draw bounding boxes around text regions as closely as possible. An example screen is shown in Figure~\ref{fig:pipeline} (a). 
To boost recall, we highlight locations of text regions found by the OCRs and workers in previous rounds. This helps workers to find initial text upon seeing the image. It also encourages workers to look for less salient text not yet detected. We first present each image once to a human worker. Subsequently, we select images that are most likely to contain further text and show them to additional annotators. Workers could also use a magnifying glass to identify small text regions, which is particularly useful to annotate illegible text. We remove duplicates after transcriptions are collected to increase robustness. Annotations from this stage contributes 80\% text of all regions and in particular 96\% of illegilbe text. 

\begin{figure*}[t]
	\includegraphics[width=1\linewidth]{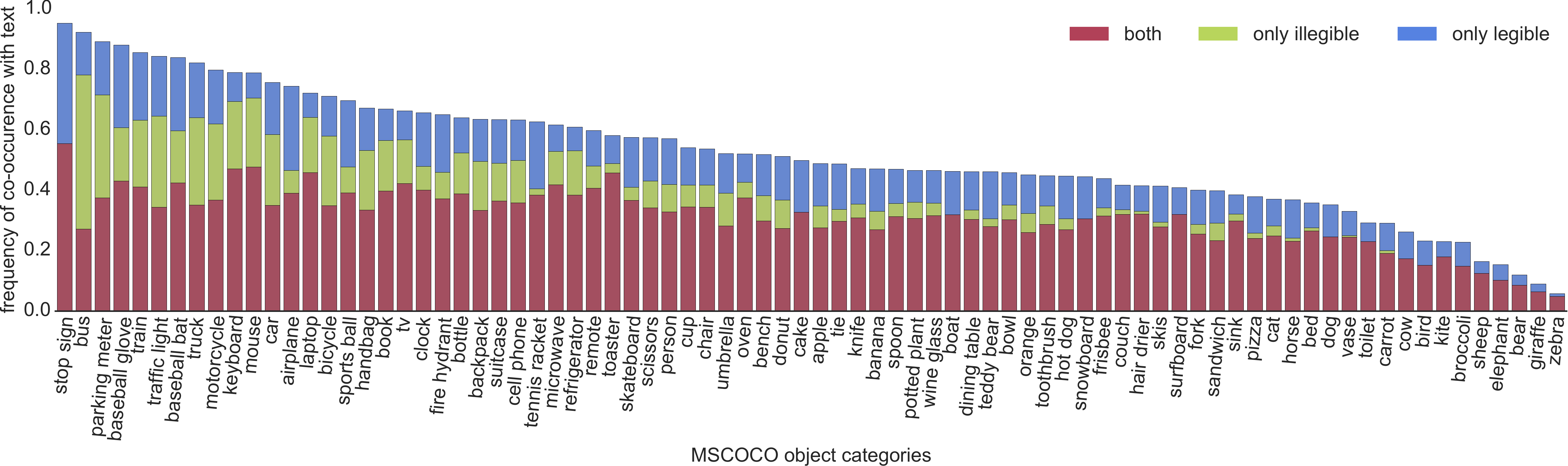}
	\caption{
		Frequency that objects in MS COCO co-occur with text. It can be seen that the presence of certain objects is very informative regarding text presence. Especially traffic and sports scenes almost always contain text and nature scenes with animals rarely contain text.
	}
	\label{fig:frequency}
	\vspace{-11pt}
\end{figure*}

\textbf{Images that need more attention\ } 
After each image has been annotated by one worker, we select images that need additional attention by workers. We use four decision rules and show images that need the most attention to workers first. We filter images whose annotators in the first round found significantly less text than the average worker. Further, we select images, in which many text regions were found in the first round. Complex natural scenes contain large amounts of text, however, workers rarely annotate all of them. For the next two rules we use the MS COCO object categories present in the image. We select images containing objects frequently having text on them, but during initial annotation no text was labeled. Lastly, we select images containing objects that tend to be in images where workers often find annotations in subsequent rounds. To learn the parameters we select 1000 images with balanced object frequencies and annotate them with high redundancy.

\textbf{Removing false positives\ } After collecting all detections we remove false positives. We use grids of crops around text regions and workers select those that do not contain text. An analogous interface is shown in Figure~\ref{fig:pipeline}~(b). We present each region proposal to three workers and use the majority rule to decide which detections are false.  

\subsection{Fine-Grained Text Classification}
Next, we classify detected text regions according to three attributes: First, legibility in terms of \emph{legible} and \emph{illegible} text. Legibility can be understood as indicator whether text can be read. Second, we collect the script of the text. We separate text into \emph{English} and \emph{not English}. We first aimed to separate all text by language. However, almost all text is English and the language of names, e.g., from restaurants, can be very ambiguous. Consequently we grouped all text that is of western script as \emph{English}, although it contains very small fractions of German, French and Spanish text. We grouped all remaining text into \emph{not English}. Third, we classify the type of text into \emph{machine printed}, \emph{handwritten} and \emph{others}. The latter included borderline text that cannot be classified as well as categories not individually covered. 

Each attribute requires a slightly different approach for the annotation. For legibility, we aim for high recall of legible text regions, because they go to the subsequent transcription stage. Using a grid interface as shown in Figure~\ref{fig:pipeline}~(b), workers select all legible text regions. We repeat this three times and each time remove those already selected. Selected text that turns out illegible will be identified in the transcription stage. To separate the script, we only use text marked as legible and use a majority vote of three annotators to select text that is not English or numerical. To separate machine printed from handwritten text, we first take a majority vote of three workers to identify handwritten text. Subsequently we identify machine printed text in another majority vote of the remaining text instances. Text not identified as either is categorized as \emph{others}. 

\small
\begin{table*}[t]
	\centering
	\caption[Caption for LOF]{State-of-the-art photo OCR detection, transcription and end-to-end results on COCO-Text. Results are on the validation set with 20.000 images and shown in percent. Detection is evaluated for all types of text and broken down into sub-categories. Text recognition and End-to-end performance is only evaluated on legible machine printed and handwritten text of English script. (Results computed with Evaluation API v.1.3\footnotemark)}
	\label{tab:OCRresults}
	\begin{tabular}{lrrrrrrrrrrr}
		\hline
		\multicolumn{1}{l|}{Alg} &  \multicolumn{7}{c|}{Localization} &  \multicolumn{1}{c|}{Recognition}  & \multicolumn{3}{c}{End-to-end}  \\ \hline
		
		\multicolumn{1}{l|}{}             &  \multicolumn{5}{c|}{recall}        & \multicolumn{1}{c|}{precision}& \multicolumn{1}{c|}{f-score} & \multicolumn{1}{c|}{accuracy}  & \multicolumn{1}{c}{recall} & \multicolumn{1}{c}{precision}&\multicolumn{1}{c}{f-score} \\ \cline{2-8}
		
		\multicolumn{1}{l|}{}             &  \multicolumn{2}{c|}{legible}  &  \multicolumn{2}{c|}{illegible}  &  \multicolumn{1}{c|}{total} & \multicolumn{1}{c|}{total}     & \multicolumn{1}{c|}{total} & \multicolumn{1}{c|}{}   \\ \cline{2-5}
		
		\multicolumn{1}{l|}{}           & \multicolumn{1}{r}{machine} & \multicolumn{1}{r|}{hand} & \multicolumn{1}{r}{machine} & \multicolumn{1}{r|}{hand} & \multicolumn{1}{l|}{}  & \multicolumn{1}{l|}{}           &\multicolumn{1}{l|}{}           & \multicolumn{1}{l|}{}  \\ \cline{1-12}
		
        A     & 34.01  &  15.11  &  4.09  &  2.31 & 23.3  &  83.78  &  36.48  &  82.91  &  28.33  &  68.42  &  40.07\\
        B     & 16.16  &  10.96  &  0.69  &  0.22 & 10.7  &  89.73  &  19.14  &  61.01  &  9.97  &  54.46  &  16.85\\
        C     & 7.09  &  4.63  &  0.28  &  0.22 & 4.7  &  18.56  &  7.47  &  23.42  &  1.66  &  4.15  &  2.37\\  \hline
	\end{tabular}
	\vspace{-10pt}
\end{table*}
\normalsize

\subsection{Character Recognition}
Lastly, we collect the transcriptions. The collection has three iterations and each consists of two steps. First we collect transcriptions and second we check with majority vote whether they are correct. In the first iteration we take transcriptions provided by the OCRs and ask workers to check for correctness. In the second and third iteration human annotators transcribe and check the text regions. The interface as shown in Figure~\ref{fig:pipeline}~(c) presents crops of text regions and the whole image for context. Worker can flag the text as illegible if they cannot read it. In all iterations we present the text regions marked as incorrectly transcribed or illegible in previous iterations. We keep transcriptions marked as correct in each stage and annotate those marked illegible and incorrectly transcribed in the last iteration as illegible text.

\subsection{Annotation Performance Analysis}
To evaluate the performance of our annotation pipeline we collected ground truth annotations for a random subset of 1000 images. We used expert annotators (co-authors of the paper) to accurately annotate all text regions. We observe that the crowd annotators detect 57\% of all text regions. In particular, they detect 84\% of legible and 39\% of illegible text. This result illustrates how difficult text detection on the COCO dataset is even for human annotators. Among the detected text regions we also analyze the classification and transcription performance. We observe the classification into machine printed and handwritten text is 93\% accurate. For borderline text, crowd annotators tend towards \emph{others}, whereas expert annotators tend towards \emph{machine printed}. Regarding text legibility, we observe an accuracy of 87\%. For borderline text, crowd annotators tend towards \emph{illegible}, whereas expert annotators slightly tend towards \emph{legible}. For script we observe an accuracy of 99\%. Crowd transcriptions are identical to the ground truth for 87.5\% of text regions and 92.5\% are within an edit distance of 1, which includes mainly missing punctuation.
\section{Dataset Statistics}\footnotetext{Evaluation API is availabe at https://github.com/andreasveit/coco-text}
We analyze COCO-Text and compare it to other popular scene text datasets, in particular, with ICDAR 03 \cite{lucas2003icdar} and ICDAR 15 \cite{karatzas2015icdar}. Previous related datasets were specifically collected for detecting and recognizing scene text. COCO-Text is based on MSCOCO, which is designed for detecting and segmenting objects occurring in their natural context. This leads to three key differences between COCO-Text and related datasets. 
First, the images of COCO-Text were not collected with text in mind. This leads to a higher variety of text and generally more natural text. As a consequence, and shown in Figure~\ref{fig:datasets}~(e), the spatial text occurrence probability, as in \cite{karaoglu2012object}, is wider distributed in COCO-Text than in related datasets. Further, COCO-Text is the only scene text datset containing images without any text. As shown in Figure~\ref{fig:datasets}~(c), 50\% of the images do not contain text. It is worthy of note that half of the images contain some form of text, although they have not been collected with text in mind. Overall, there are in average $2.73$ instances of text per image. Considering only images with text, the average is $5.46$.
An important property of this dataset is that, text instances are annotated with more attributes than in related datasets. This is especially useful due to the high variety of text in natural images. Figure~\ref{fig:datasets}~(b) gives an overview of how many text attributes related datasets contain. In addition to location, traditional datasets only contain transcriptions. ICDAR 15 also includes ``do not care'' regions that could be interpreted as a notion of text legibility. COCO-Text goes further by also annotating the type of the text, \ie whether it is machine printed or handwritten, as well as the script of the text. The number of instances per category is shown in Figure~\ref{fig:datasets}~(a). Overall, 60.3\% of text is legible and 39.7\% illegible. The majority of text is machine printed with 86.4\%. Only 4.6\% of text is handwritten and 9\% is borderline or from other not captured categories.
Another key aspect of COCO-Text is its contextual information. It is part of the larger context of MSCOCO and thus enables contextual reasoning between scene text and objects. This is relevant as context is highly informative for many applications. Figure~\ref{fig:frequency} shows the frequency of co-occurrences of MSCOCO object categories with scene text. It can be seen that the presence of certain objects is very informative regarding text presence. 
Lastly, COCO-Text has a larger scale than related datasets. Containing 63,686 images and 173.589 text annotation, it is more than 14 times larger than the latest ICDAR 15. Figure~\ref{fig:datasets}~(b) provides an overview.

\section{Algorithmic Analysis}
In this section we evaluate the current state-of-the-art in photo OCR on our dataset. Using the wide variety of text and annotations, we are interested in identifying areas where the performance is particular strong and areas with significant shortcomings motivating future work.

\textbf{Evaluation procedure\ } 
We follow the text detection and recognition evaluation scheme as used in the ICDAR robust reading competition for end-to-end recognition of \emph{incidental scene text}~\cite{karatzas2015icdar}. The evaluations for scene text detection uses a single Intersection-over-Union (IoU) criterion, with a threshold of 50\%, similar to object detection \cite{everingham2014pascal}. If multiple detection bounding boxes satisfy the threshold for a ground truth box, the best match is identified as the one with correct text recognition and otherwise with the highest IoU score. Text is annotated with one bounding box per word, \ie an uninterrupted sequence of characters separated by a space. For end-to-end results we only consider a detection a correct match if the words match~\cite{wang2011end}. Recognition and End-to-end performance is only evaluated on legible machine printed and handwritten text of English script.

\textbf{Evaluation results\ } In our experiments we take three state-of-the-art photo OCR algorithms from our collaborators at Google, TextSpotter and VGG and evaluate their detection, transcription and end-to-end text spotting results on our dataset. Since this is not a competition, they are anonymized in Table~\ref{tab:OCRresults}, which shows the evaluation results. On the positive side, methods A and B have good detection precision with 83.78 and 89.73\% respectively. Further, we observe good recognition accuracy. In particular, method A achieves 82.91\%. However, detection performance is very weak overall. While method A finds considerable amounts of legible machine printed text with 34.01\%, no method performs satisfactory. Even lower results are observed on legible handwritten text. These unsatisfactory detection results on natural images in COCO-Text motivate future work. Lastly, no method has even viable functionality to find illegible text. It is worthy of note that current photo OCR algorithms are not supposed to detect or transcribe illegible text. As a consequence, it requires novel methods to fill this research gap. Note that these approaches are used in our annotation and although we ensure redundancy with human annotators the results are not a baseline.
\section{Dataset Split}
The dataset is split into training and validation set, which contain 43686 and 20000 images respectively. To report end-to-end text spotting results only legible machine printed and handwritten text should be considered. We encourage researchers to train and adjust parameters on the training set, but minimize the number of runs on the evaluation set.
\section{Discussion}
We introduced COCO-Text\footnote{available at http://vision.cornell.edu/se3/coco-text} a new dataset for detecting and recognizing text in natural images to support the advancement of text recognition in everyday life environments. Using over 1500 worker hours, we annotated a large collection of text instances spanning several types of text. This is the first large-scale dataset for text in natural images and also the first dataset to annotate scene text with attributes such as legibility and type of text. Dataset statistics indicate the images contain a wide variety of text and the spatial distribution of text is broader than in related datasets. We further evaluate state-of-the-art photo OCR algorithms on our dataset. While the results indicate satisfactory precision, we identify significant shortcomings especially for detection recall. This motivates future work towards algorithms that can detect wider varieties of text. We believe this dataset will be a valuable resource supporting this effort.
\section*{Acknowledgments}
Funding for the crowd worker tasks was provided by a Microsoft Research Award. We would like to thank our collaborators Kai Wang from Google 
and Ankush Gupta and Andrew Zisserman from Oxford for providing photo OCR output for our data collection effort and for providing valuable feedback throughout the process of defining and collecting the dataset. A.V. was supported by the AOL-Program for Connected Experiences.

{\small
\bibliographystyle{ieee}
\bibliography{egbib}

\begin{thebibliography}{10}\itemsep=-1pt

\bibitem{bissacco2013photoocr}
A.~Bissacco, M.~Cummins, Y.~Netzer, and H.~Neven.
\newblock {PhotoOCR: Reading text in uncontrolled conditions}.
\newblock In {\em ICCV '13}, pages 785--792. IEEE, 2013.

\bibitem{chen2015microsoft}
X.~Chen, H.~Fang, T.-Y. Lin, R.~Vedantam, S.~Gupta, P.~Dollar, and C.~L.
  Zitnick.
\newblock Microsoft coco captions: Data collection and evaluation server.
\newblock {\em arXiv preprint arXiv:1504.00325}, 2015.

\bibitem{deng2009imagenet}
J.~Deng, W.~Dong, R.~Socher, L.-J. Li, K.~Li, and L.~Fei-Fei.
\newblock Imagenet: A large-scale hierarchical image database.
\newblock In {\em CVPR '09}, pages 248--255. IEEE, 2009.

\bibitem{everingham2014pascal}
M.~Everingham, S.~A. Eslami, L.~Van~Gool, C.~K. Williams, J.~Winn, and
  A.~Zisserman.
\newblock The pascal visual object classes challenge: A retrospective.
\newblock {\em International Journal of Computer Vision (IJCV)},
  111(1):98--136, 2014.

\bibitem{jaderberg2014synthetic}
M.~Jaderberg, K.~Simonyan, A.~Vedaldi, and A.~Zisserman.
\newblock Synthetic data and artificial neural networks for natural scene text
  recognition.
\newblock {\em arXiv preprint arXiv:1406.2227}, 2014.

\bibitem{jaderberg2014reading}
M.~Jaderberg, K.~Simonyan, A.~Vedaldi, and A.~Zisserman.
\newblock Reading text in the wild with convolutional neural networks.
\newblock {\em International Journal on Computer Vision (IJCV)}, pages 1--20,
  2015.

\bibitem{Karaoglu13}
S.~Karaoglu, J.~van Gemert, and T.~Gevers.
\newblock Con-text: text detection using background connectivity for
  fine-grained object classification.
\newblock In {\em {ACM Multimedia (ACM MM)}}, 2013.

\bibitem{karaoglu2012object}
S.~Karaoglu, J.~C. Van~Gemert, and T.~Gevers.
\newblock Object reading: text recognition for object recognition.
\newblock In {\em ECCV '12. Workshops and Demonstrations}, pages 456--465.
  Springer, 2012.

\bibitem{karatzas2015icdar}
D.~Karatzas, L.~Gomez-Bigorda, A.~Nicolaou, S.~Ghosh, A.~Bagdanov, M.~Iwamura,
  J.~Matas, L.~Neumann, V.~R. Chandrasekhar, S.~Lu, et~al.
\newblock Icdar 2015 competition on robust reading.
\newblock In {\em Document Analysis and Recognition (ICDAR), 2015 13th
  International Conference on}, pages 1156--1160. IEEE, 2015.

\bibitem{lin2014microsoft}
T.-Y. Lin, M.~Maire, S.~Belongie, J.~Hays, P.~Perona, D.~Ramanan,
  P.~Doll{\'a}r, and C.~L. Zitnick.
\newblock Microsoft coco: Common objects in context.
\newblock In {\em ECCV '14}, pages 740--755. Springer, 2014.

\bibitem{lucas2003icdar}
S.~Lucas, A.~Panaretos, L.~Sosa, A.~Tang, S.~Wong, and R.~Young.
\newblock Icdar 2003 robust reading competitions.
\newblock In {\em ICDAR '03}, pages 682--687. IEEE, 2003.

\bibitem{matera2014user}
T.~Matera, J.~Jakes, M.~Cheng, and S.~Belongie.
\newblock A user friendly crowdsourcing task manager.
\newblock In {\em Workshop on Computer Vision and Human Computation (CVPR)},
  2014.

\bibitem{mishra2012scene}
A.~Mishra, K.~Alahari, and C.~Jawahar.
\newblock Scene text recognition using higher order language priors.
\newblock In {\em BMVC '12}. BMVA, 2012.

\bibitem{neumann2012real}
L.~Neumann and J.~Matas.
\newblock Real-time scene text localization and recognition.
\newblock In {\em CVPR '12}, pages 3538--3545. IEEE, 2012.

\bibitem{romney1986culture}
A.~K. Romney, S.~C. Weller, and W.~H. Batchelder.
\newblock Culture as consensus: A theory of culture and informant accuracy.
\newblock {\em American anthropologist}, 88(2):313--338, 1986.

\bibitem{rusinol2014multimodal}
M.~Rusi{\~n}ol, V.~Frinken, D.~Karatzas, A.~D. Bagdanov, and J.~Llad{\'o}s.
\newblock Multimodal page classification in administrative document image
  streams.
\newblock {\em International Journal on Document Analysis and Recognition
  (IJDAR)}, 17(4):331--341, 2014.

\bibitem{RussakovskyCVPR15}
O.~Russakovsky, L.-J. Li, and L.~Fei-Fei.
\newblock Best of both worlds: human-machine collaboration for object
  annotation.
\newblock In {\em CVPR '15}, 2015.

\bibitem{veit2015optimizing}
A.~Veit, M.~Wilber, R.~Vaish, S.~Belongie, J.~Davis, et~al.
\newblock On optimizing human-machine task assignments.
\newblock In {\em AAAI Conference on Human Computation and Crowdsourcing
  (HCOMP)}, San Diego, CA, 2015.
\newblock WiP.

\bibitem{wang2011end}
K.~Wang, B.~Babenko, and S.~Belongie.
\newblock End-to-end scene text recognition.
\newblock In {\em ICCV '11}, pages 1457--1464. IEEE, 2011.

\bibitem{wang2010word}
K.~Wang and S.~Belongie.
\newblock Word spotting in the wild.
\newblock In {\em ECCV '10}.

\bibitem{wilber2014cost}
M.~J. Wilber, I.~S. Kwak, and S.~J. Belongie.
\newblock Cost-effective hits for relative similarity comparisons.
\newblock In {\em AAAI Conference on Human Computation and Crowdsourcing
  (HCOMP)}, 2014.

\bibitem{xiao2010sun}
J.~Xiao, J.~Hays, K.~Ehinger, A.~Oliva, A.~Torralba, et~al.
\newblock Sun database: Large-scale scene recognition from abbey to zoo.
\newblock In {\em CVPR '10}, pages 3485--3492. IEEE, 2010.

\end{thebibliography}
}

\end{document}